%% file: main.tex
\documentclass{article}
\PassOptionsToPackage{numbers,sort&compress}{natbib}
\usepackage[sglblindworkshop,final]{neurips_2026}
\workshoptitle{AI at Scale for Clinical Impact: Cancer Pathology Foundation Models}

\usepackage[utf8]{inputenc}
\usepackage[T1]{fontenc}
\usepackage{booktabs}
\usepackage{graphicx}
\usepackage[hidelinks]{hyperref}
\usepackage{microtype}
\usepackage{tabularx}
\usepackage{url}
\usepackage{xcolor}
\usepackage{colortbl}
\usepackage{listings}
\usepackage{amsmath}

\title{Nanopath: Fast, Fair, Open Experimentation for Pathology Foundation Models}

\author{%
  \textbf{Paul S. Scotti}$^{1,2}$ \quad \textbf{Hassan Muhammad}$^{1,2}$ \quad \textbf{Ryan Kim}$^{1,3}$ \\
  \textbf{Nima Ashjaee}$^{1,4}$ \quad \textbf{Ratna Sagari Grandhi}$^1$ \quad \textbf{Saman Rahbar}$^{1,4}$ \\
  \textbf{Raymond Biju}$^1$ \quad \textbf{Anish Dulal}$^{1,5}$ \quad \textbf{Chao Feng}$^1$ \\
  \textbf{Connor Lane}$^{1,2}$ \quad \textbf{Benjamin Warner}$^{1,2}$ \quad \textbf{Tanishq Mathew Abraham}$^{1,2}$ \\[3pt]
  $^1$MedARC \quad $^2$Sophont \\
  $^3$University of Illinois at Urbana-Champaign \\
  $^4$University of British Columbia \quad $^5$University of Oregon \\
  \texttt{paul@sophont.med} \\
  \texttt{https://medarc.ai}
  \texttt{ | https://labless.dev}
}

\begin{document}

\maketitle

\begin{abstract}
We introduce nanopath, an open-source framework and benchmark for rapidly training and evaluating small-scale pathology foundation models. Using a single H100 GPU, a typical nanopath model trains in a little over an hour on up to one million TCGA pathology tile presentations and evaluates within 25 minutes on a broad suite of tile-level and slide-level tasks. Nanopath enables rapid exploration of research ideas at small scale, so promising training recipes can subsequently be tested with more data and compute. To enable fair comparisons between ideas, we constrain submissions to fixed data and compute budgets while keeping the downstream evaluation protocol fixed. Anyone may clone the codebase, automatically download the data, reproduce or modify existing models, and submit a run to Labless, our public experiment ledger where researchers and AI agents can search through prior attempts. Since its public release in June 2026, nanopath has received 1,269 experiments from 20 contributors. Here, we detail its design and assess how well its rapid evaluation tracks standardized THUNDER, HEST, and PathoBench (CPTAC) benchmarks across six nanopath and 14 reference foundation models. Nanopath's fast evaluation suite strongly tracked these standardized benchmarks, and a few 22M parameter nanopath models even outperformed giant 1.1B parameter models on a subset of downstream tasks. We subsequently discuss how we have since optimized nanopath's evaluation suite to be an even more faithful, quick proxy for external benchmarking. Nanopath provides an accessible testbed for students, researchers, and AI agents to explore pathology foundation modeling through fair, transparent, and directly comparable experiments.
\end{abstract}

\section{Introduction}
Pathology foundation models are increasingly developed by well-resourced academic, pharmaceutical, and commercial research teams using large proprietary datasets~\citep{chen2024uni,vorontsov2024virchow,xu2024gigapath}. This has produced strong encoders whose learned representations support a broad range of downstream tasks including tissue classification, segmentation, and molecular prediction, but relatively little controlled insight into what factors are most important to their performance. Many competitive pathology foundation models are variations on DINO-style self-supervision with different pretraining data~\citep{oquab2023dinov2,zimmermann2024virchow2}. Further, access to pretraining datasets, training code, and model weights varies, and standardized benchmark evaluations can include data sources already used during model pretraining. As a result, progress is difficult to disentangle: when one model outperforms another, it is unclear whether the gain is from differences in pretraining data, model architecture, data curation, model/compute scaling, dataset double dipping, or something else. Our earlier OpenMidnight 1.0 work demonstrated strong performance of an open-source foundation model trained in approximately 3.5 days on a single node of eight H100 GPUs using only public TCGA slides~\citep{openmidnight2025,openmidnight2026}, motivating the idea that strong performance and research insights are attainable under intense compute and time constraints.

The difficulty of directly comparing foundation models is concerning for computational pathology because it is a domain where the structure of the data leaves substantial room for methodological innovation beyond conventional natural-image datasets. Whole-slide images are gigapixel-scale, contain strong spatial and hierarchical structure, and often exhibit highly sparse biological signal: a small region of diagnostically meaningful tissue may be embedded within large amounts of normal tissue, background, artifact, or otherwise irrelevant morphology. Exploring how best to exploit this structure is time-consuming and expensive if hypotheses require training and evaluating full foundation models. A useful framework should make it possible to test many hypotheses cheaply and under controlled conditions before committing substantial data and compute to scaling.

\begin{table}[t]
\centering
\caption{Training and evaluation constraints for nanopath runs to be valid leaderboard submissions.}
\label{tab:nanopath_constraints}
\small
\setlength{\tabcolsep}{4pt}
\renewcommand{\arraystretch}{1.05}
\begin{tabularx}{\columnwidth}{@{}lX@{}}
\toprule
\textbf{Constraint} & \textbf{Requirement} \\
\midrule
Tile presentations
& $\leq 1{,}000{,}000$ \\

Training compute
& $\leq 10^{18}$ FLOPs \\

Pathology image source
& TCGA WSIs only \\

Non-pathology pretrained models
& Allowed for initialization and as teachers \\

Pathology-pretrained models
& Not allowed \\

Public clinical/genomic metadata
& Allowed \\

Pretraining preprocessing
& Excluded from the training FLOP and runtime budgets \\

Maintainer validation
& Training reproduces in $\leq 2$ hours on 80\,GB H100 with different rng seeds \\

Probe evaluation
& Fixed suite; must complete in $\leq 25$ minutes on one 80\,GB H100 \\

Evaluation protocol
& Probe code, downstream data, and configuration may not be modified \\
\bottomrule
\end{tabularx}
\end{table}

\begin{figure*}[t]
  \centering
  \includegraphics[width=1.\textwidth]{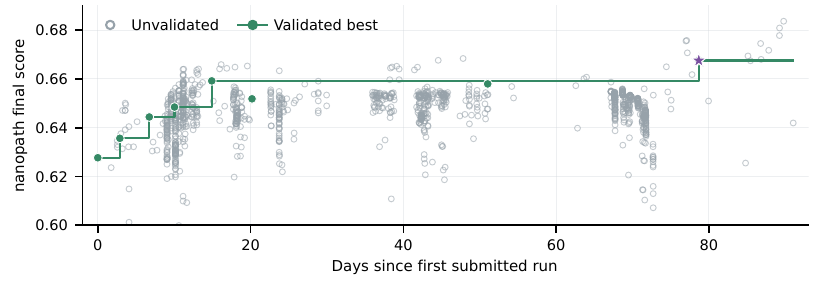}
  \caption{\textbf{Sept 1 2026 snapshot of nanopath leaderboard on Labless.} The ledger contains 1,269 runs from 20 contributors, scored with our original nanopath evaluation suite. Open points are unvalidated submissions; filled green points and purple star are maintainer-verified reproduction runs, with green curve indicating the best performing run over time. Y-axis omits low-scoring outliers.}
  \label{fig:nanopath-progress}
\end{figure*}

We developed nanopath around this principle. \href{https://github.com/MedARC-AI/nanopath}{Nanopath} provides a deliberately small and hackable experimental setting in which research ideas can be compared rapidly under common constraints. Each standard run is restricted to at most one million presentations of publicly available TCGA pathology tiles and \(10^{18}\) training FLOPs, while the downstream evaluation protocol is fixed and quick to run. Within those constraints, researchers are free to modify essentially the entire training recipe, including model architecture, learning objective, augmentations, data curation, and feature aggregation. A typical run trains on one 80\,GB H100 GPU in roughly an hour, with a two-hour ceiling for maintainer reproductions (Table~\ref{tab:nanopath_constraints}), and is then evaluated in less than 25 minutes using frozen representations across a broad suite of tile-level, slide-level, and robustness evals. The default encoder is a ViT-S/14 with approximately 22 million parameters, although the rules permit other architectures. Inspired by autoresearch's automated cycle of proposing, training, and comparing experiments~\citep{karpathy2026autoresearch}, nanopath supports both human researchers and AI agents iterating under the same data, compute, and evaluation constraints.

Nanopath is also designed as a shared collaborative leaderboard. Completed runs can be submitted directly to Labless\footnote{\url{https://labless.dev}}, our platform created specifically to support nanopath as its inaugural project. We encourage all researchers trying out nanopath to submit every completed full training run, including negative results, to Labless via our API, which shares the source code, downstream metrics, and hardware information for runs to a public record. This enables a searchable history of attempted ideas, allowing researchers and AI agents to see what has already been tried, avoid repeating the same dead ends, and identify patterns across experiments. In this sense, the goal of nanopath is not simply to reduce the cost of training a pathology foundation model, but to increase the rate at which the field can generate, falsify, share, and accumulate evidence about how pathology representations should be learned by AI models.

Such a system is only useful, however, if improvements observed through our fast evaluation suite tend to generalize to other downstream tasks. Otherwise, we could be encouraging hillclimbing on meaningless idiosyncrasies of our nanopath final score. Hence, we test nanopath's fast evaluation against THUNDER, HEST, and PathoBench (CPTAC), using six nanopath checkpoints and 14 reference encoders~\citep{marza2025thunder,jaume2024hest,zhang2025pathobench}. Overall, nanopath scores strongly correlated with scores on these external benchmarks, with some nanopath models even surpassing large-scale pathology foundation models on selected downstream metrics. We also observed weaknesses in probe selection and pretraining--evaluation data overlap, motivating us to develop an optimized evaluation suite with broader coverage and stronger agreement with standardized benchmarks.

The fact that a small 22M-parameter encoder, initialized from a natural-image model and trained briefly on public TCGA tiles, can approach much larger pathology foundation models on some downstream tasks speaks to the opportunities for experimentation in this field. Successful nanopath runs already showcase the potential of JEPA, metadata conditioning, data curation, and test-time feature aggregation. We hope nanopath will unlock further insights and make pathology foundation model research more open, accessible, and rigorous.

\section{Nanopath as a Controlled One-GPU Experiment}
\label{sec:nanopath}

\subsection{Quickstart}
\label{sec:quickstart}
A complete nanopath experiment can be reproduced in just a few lines of code:
\begin{lstlisting}[language=bash,basicstyle=\ttfamily\small,frame=single,breaklines=true,columns=fullflexible,keepspaces=true,showstringspaces=false]
git clone https://github.com/MedARC-AI/nanopath.git && cd nanopath
uv sync && source .venv/bin/activate
python prepare.py download=True
python train.py configs/main.yaml
\end{lstlisting}
This code downloads and installs all prerequisites and kicks off a full training run of a reference nanopath model, followed by subsequent evaluation of the model. The repository also provides a smoke configuration for checking installation before a full run~\citep{medarc2026nanopath}. A fresh setup requires approximately 355\,GB of free space: 120\,GB for pretraining shards, 215\,GB for extracted evaluation data, and roughly 20\,GB of temporary space during archive extraction.

\subsection{Pretraining constraints and scaffolding}
Nanopath constrains pretraining along two resource axes: data exposure and computation. Each run is configured with a maximum of 1M tile presentations and \(10^{18}\) training FLOPs (Table~\ref{tab:nanopath_constraints}). A tile presentation corresponds to one TCGA pathology tile emitted by the dataloader; repeated exposure to the same underlying tile counts as another presentation, whereas multiple crops, masks, or teacher/student views derived from a single emission collectively count as one presentation. Their associated computation, however, contributes to the FLOP budget. The implementation uses PyTorch's operator FLOP counter over the first complete training step, including forward and backward passes, optimizer updates, and teacher networks, and reuses that measured cost for subsequent fixed-shape steps. 

The image data available for pretraining are restricted to TCGA whole-slide images~\citep{weinstein2013tcga}. Researchers are free to modify tile extraction, filtering, sampling, curation, preprocessing, model architecture, training objective, optimization, and augmentation strategies within this constraint. Public clinical or genomic metadata, annotations, text, and ontologies may also be incorporated. Distillation from a pathology foundation model during training and pathology-pretrained initialization are disallowed. Non-pathology pretrained models, such as DINOv2, may be used for initialization, as teachers, or during preprocessing. Computationally intensive preprocessing is excluded from the training budget, so inherited models and preprocessing costs must be disclosed for meaningful comparison.

Nanopath distributes four million $224\!\times\!224$ pixel tiles from 11,368 TCGA diagnostic whole-slide images, randomizing microns-per-pixel (MPP) resolution following OpenMidnight's preprocessing approach~\citep{openmidnight2025,openmidnight2026}. All slides are formalin-fixed, paraffin-embedded (FFPE). The dataset is provided as 200 Arrow shards totaling roughly 120\,GB, allowing researchers to begin training without downloading or processing the approximately 13\,TB of original TCGA slides. Tiles were sampled from tissue-containing regions, extracted with OpenSlide~\citep{goode2013openslide}, encoded as JPEG images, and packaged for efficient random access during training. Researchers may use these pre-extracted tiles directly or regenerate the dataset themselves. Alternative pipelines must include reproducible code to qualify for leaderboard validation.

\subsection{A fast proxy for broad downstream evaluation: nanopath-evals}
\label{sec:eval}
A central design goal of nanopath's fast evaluation suite (termed \texttt{nanopath-evals}) is to combine broad downstream coverage with a short runtime. After pretraining, we freeze the encoder and evaluate its representations across a fixed collection of lightweight downstream probes. The suite covers tile classification, semantic segmentation, slide-level progression, mutation prediction, survival prediction, and representation robustness. Its protocols are adapted from THUNDER, PathoBench, LEOPARD, PathoROB, and CRoMa, with modifications that make repeated evaluation practical on a single GPU~\citep{marza2025thunder,zhang2025pathobench,faryna2026leopard,komen2026pathorob,grisi2026croma}. \texttt{nanopath-evals} must finish within 25 minutes on one 80\,GB H100 GPU, whereas the external THUNDER, HEST, and PathoBench evaluations require multiple GPU and CPU stages, with individual stages taking minutes to hours (Appendix~\ref{app:external}).

All models use identical probe implementations and fixed data splits. The encoder is never fine-tuned: classification and slide-level probes operate on frozen features, segmentation trains only a lightweight decoder, and robustness uses a non-parametric neighborhood analysis. The fast probes draw from official development splits or training pools. We do not include any test samples from corresponding official benchmarks in \texttt{nanopath-evals}. This preserves test data for subsequent external evaluation. Appendix~\ref{app:protocol} details our data selections and splits.

\paragraph{Efficient single-GPU evaluation.}
Most of the efficiency behind our evaluation suite comes from embedding small, representative data subsets in half-precision batches and reusing frozen features across probe methods, hyperparameter settings, and few-shot trials.

For classification, nanopath uses fixed class-, source-, and group-aware subsets from the same underlying data sources as THUNDER, while holding out THUNDER's test splits. Linear heads share cached image features, train jointly across the hyperparameter grid, and are scored only at the final epoch. All $k$-NN settings reuse one chunked GPU similarity search. SimpleShot evaluates only the 16-shot setting, with all 1,000 support draws processed in GPU batches~\citep{wang2019simpleshot}. These choices avoid repeated backbone inference, validation passes, neighbor sorting, and data loading overhead.

Segmentation, slide-level, and robustness evaluations follow the same frozen-feature design. For segmentation, a compact THUNDER-style MaskTransformer trains on cached int8 patch features using a Dice objective~\citep{strudel2021segmenter}. Expanded spatial readouts are pooled to the native patch grid while retaining all feature channels. For slide-level tasks, pre-tiled images are embedded in large 16-bit batches and mean-pooled online into one vector per slide or patient. UCLA Lung and CPTAC-PDA use their complete prepared tile grids; SurGen and LEOPARD use at most 768 spatially distributed tiles per slide. Progression and mutation use logistic regression, while survival uses CoxNet (an elastic-net-regularized Cox proportional hazards model)~\citep{simon2011coxnet}. Finally, PathoROB images are embedded once using the fixed CLS-plus-mean-patch representation before chunked neighborhood analysis.

\paragraph{Original and optimized suites.}
The original nanopath evaluation suite averaged eight terms: linear, $k$-NN, 16-shot, segmentation, progression, mutation, survival, and robustness (Appendix~\ref{app:original}). Classification therefore received three-eighths of the score despite covering only four datasets. As of Sept. 1 2026, we optimized the suite to add more evaluation datasets, exclude fully TCGA-derived tasks, more closely align individual probes with the corresponding external benchmarks, and reweight the final score. Classification expands from four datasets to 12 (Table~\ref{tab:nanopath_eval_v2}). Rather than select the best linear-probe or $k$-NN hyperparameter on the same validation samples used for scoring, we average performance across THUNDER's fixed grids. Segmentation replaces MoNuSAC and CoNSeP with SegPath epithelial and lymphocyte tasks~\citep{komura2023segpath}, adopts weighted per-image macro-F1 instead of macro-Jaccard, and uses fixed training schedules. Slide-level datasets remain unchanged, with fixed logistic-regression heads and fold-standardized CoxNet heads at three relative regularization strengths. Finally, we replaced PathoROB's robustness index with CRoMa, which uses common sample support and relative distances to address limitations of the original index~\citep{grisi2026croma}.

\begin{table}[t]
\centering
\caption{The optimized \texttt{nanopath-evals} suite. Weight denotes contribution to nanopath final score. AUROC is macro one-versus-rest; survival uses Harrell's c-index. See Table~\ref{tab:nanopath_eval_v1} for details on the non-optimized, original nanopath suite.}
\label{tab:nanopath_eval_v2}
\small
\setlength{\tabcolsep}{4pt}
\begin{tabularx}{\textwidth}{@{}lXlr@{}}
\toprule
Family & Datasets & Metric & Weight \\
\midrule
Classification & BACH, BRACS, BreaKHis, CRC, ESCA, MHIST, PCam, SPIDER Breast/CRC/Skin/Thorax, WILDS & Macro-F1 & 30\% \\
Segmentation & PanNuke, SegPath epithelial/lymphocytes & Image-weighted F1 & 15\% \\
Progression & UCLA Lung (90 slides) & AUROC & 17.5\% \\
Mutation & SurGen RAS (311 slides) & AUROC & 15\% \\
Survival & LEOPARD BCR (174 cases), CPTAC-PDA OS (77 cases) & c-index & 7.5\% \\
Robustness & PathoROB Camelyon, non-TCGA Tolkach ESCA & CRoMa & 15\% \\
\bottomrule
\end{tabularx}
\end{table}

\paragraph{Robustness and score weighting.}
CRoMa measures whether biological similarity dominates acquisition-center similarity~\citep{grisi2026croma}. It compares biological and acquisition-center neighborhoods using relative distances, excluding same-slide neighbors. We map cohort-level median margins to $[0,1]$ and average the two cohorts to obtain $R$ (Appendix~\ref{app:protocol}). The final score is
\begin{equation}
S = 0.30C + 0.15\mathrm{Seg} + 0.175P + 0.15M + 0.075\mathrm{Sur} + 0.15R,
\label{eq:score}
\end{equation}
where $S$ is the optimized final score and $C$ averages the three classification probe types within each dataset and then averages datasets. $\mathrm{Seg}$ is mean segmentation F1; $P$ and $M$ are progression and mutation AUROC; $\mathrm{Sur}$ is mean survival c-index; and $R$ is mean rescaled CRoMa. All family means weight their datasets equally.

\subsection{Submission, validation, and integration with Labless}
\label{sec:labless}
Nanopath integrates with our platform Labless to provide a public, source-linked record of training experiments. After training and evaluation finish, the submission client reads the metrics, hardware metadata, and source snapshot captured at launch, constructs a reviewable artifact, and submits the record through the Labless API. A verified GitHub account preserves contributor attribution. The run then appears automatically in the public experiment log and progress plot without requiring a pull request. The submission pipeline audits resource caps and the locked evaluation code, and checks consistency between the final score and component metrics. These checks catch explicit rule violations and support source inspection; promising runs then undergo independent reproduction by nanopath maintainers who retrain and evaluate the submissions via different random seeds.

We encourage all valid nanopath runs to be submitted, regardless of their performance (crashed, incomplete, and smoke runs are not valid submissions). The public ledger can therefore be searched by humans and AI agents to reveal insights surrounding successful and unsuccessful recipe choices. Nanopath is therefore inherently structured to support autoresearch-style loops that propose, run, and compare experiments~\citep{karpathy2026autoresearch}.

\section{Results}
\label{sec:results}
\subsection{Progression of nanopath models}
The starting recipe, \texttt{dinov2-s-kde}, uses a compact DINOv2-S/14 implementation with four register tokens, initialized from Meta's natural-image checkpoint~\citep{oquab2023dinov2,darcet2024registers}. It retains DINO/iBOT self-distillation~\citep{caron2021dino,zhou2022ibot} and replaces KoLeo with KDE uniformity regularization, following pathology recipes such as Virchow2 and Midnight~\citep{zimmermann2024virchow2,karasikov2025midnight}. The latent-prediction recipes draw on I-JEPA~\citep{assran2023ijepa}, replacing iBOT for prediction of patch-level token features. All six checkpoints studied here use the same approximately 22M-parameter backbone, although their training and feature readouts differ. Table~\ref{tab:validated-lineage} describes the selected recipe history and Figure~\ref{fig:nanopath-progress} depicts them on an autoresearch-style plot. 

\begin{table}[t]
\centering
\caption{Six nanopath checkpoints in chronological order. Scores show original nanopath evaluation scores; $\Delta$ compares adjacent rows. Minutes are the recorded training wall-clock times, excluding evaluation probing. $\dagger$ denotes a rerun that did not clear our leader-promotion margin of .006. Exact run identifiers are in Appendix~\ref{app:lineage}.}
\label{tab:validated-lineage}
\small
\setlength{\tabcolsep}{3pt}
\begin{tabularx}{\textwidth}{@{}p{0.23\textwidth}rrrX@{}}
\toprule
Recipe & Score & $\Delta$ & Min. & Main change \\
\midrule
\texttt{dinov2-s-kde} & .6277 & -- & 73.0 & DINO/iBOT on TCGA with KDE uniformity. \\
\texttt{lr-and-curation} & .6357 & +.0080 & 59.2 & Skips tiles with $>50\%$ background, shortens warmup to 30,000 tile presentations, and sets AdamW $\beta_2=.99$. \\
\texttt{i-jepa-contig-patch} & .6444 & +.0088 & 57.9 & Smooth-L1 prediction of teacher features from contiguous target blocks replaces iBOT; gentler stain jitter and larger local crops. \\
\texttt{jepa-fino}$^\dagger$ & .6485 & +.0041 & 65.0 & Subtype, expression, and genome-alteration metadata guidance~\citep{gardes2026fino}; learning-rate and jitter changes. \\
\texttt{block-strided-cls} & .6592 & +.0108 & 54.4 & Concatenates normalized CLS features from blocks 4, 6, 8, and 11; fuses the last four blocks and applies joint bilateral upsampling for dense segmentation features. \\
\texttt{robust-norm} & .6676 & +.0083 & 100.7 & More/larger local crops, stronger KDE, and an additional CLS feature from block 2; removes feature directions induced by gamma, gain, hue, and saturation perturbations. \\
\bottomrule
\end{tabularx}
\end{table}

\subsection{External evaluation on THUNDER, HEST, and PathoBench (CPTAC)}
\label{sec:external-validation}
Nanopath models hillclimb on a final score that aggregates performance across our fast evaluation suite. This score is only useful if improvements also tend to correspond to improvements under slower, more comprehensive external evaluations. We therefore evaluated six nanopath models and 14 reference models (including 22M-parameter natural-image models and giant one billion parameter state-of-the-art pathology models). Each encoder was frozen and used its documented preprocessing and feature readout. We used already published benchmark results whenever possible, and otherwise conducted our own benchmarking using slightly adapted versions of the corresponding official benchmark codebases (Appendix~\ref{app:external}). 

THUNDER covers tile classification and segmentation~\citep{marza2025thunder}. We average its linear, $k$-NN, and 16-shot aggregates for classification, average its four segmentation datasets, and weight these two families equally for the combined target. HEST evaluates spatial gene-expression prediction on nine cancer--organ datasets using PCA-256 features and ridge regression; its task metric is correlation between predicted and measured gene expression~\citep{jaume2024hest}. The CPTAC target averages macro one-versus-rest AUROC across 38 patient-level classification tasks from 11 cohorts using mean-pooled slide representations and PathoBench protocols~\citep{zhang2025pathobench}.

\begin{figure}[!t]
\centering
\includegraphics[width=\textwidth]{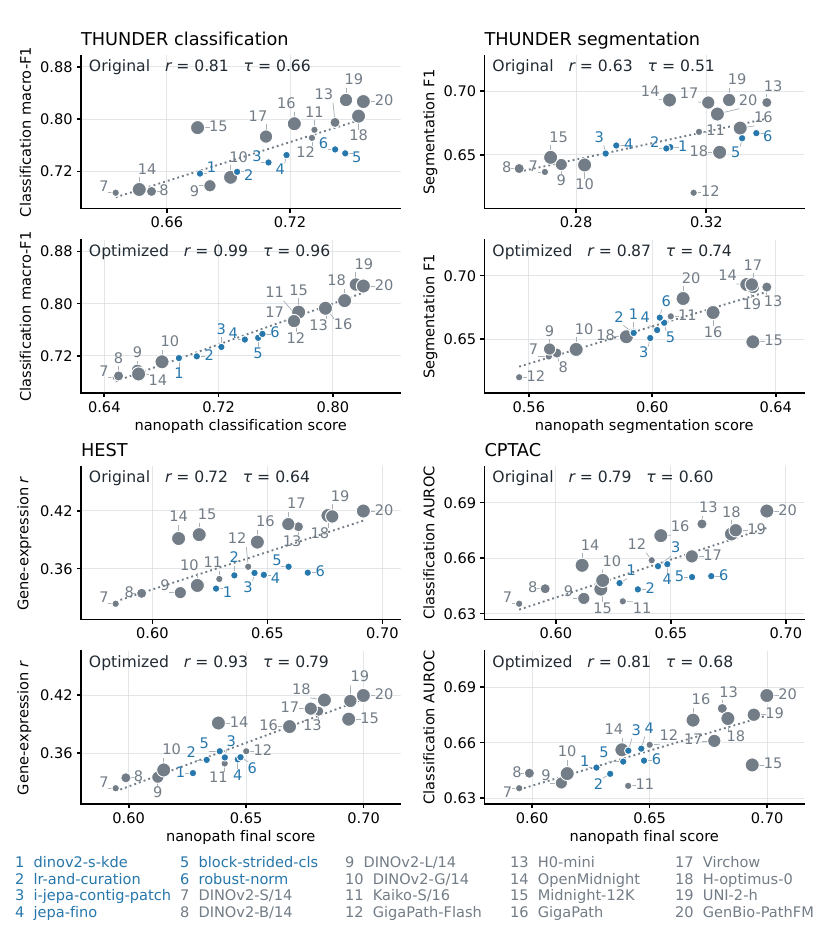}
\caption{\textbf{Original and optimized nanopath evaluations compared with external benchmarks.} Each panel contains the same 20 frozen encoders, identified by the shared key. Blue denotes nanopath models; marker area is log-scaled by parameter count (from 22M to 1B parameters). $r$ is Pearson correlation and $\tau$ Kendall rank correlation. THUNDER panels compare matching family scores; HEST and CPTAC use the nanopath final score. See Appendix Table~\ref{tab:allmodels} for exact metrics. OpenMidnight refers to v1.0, our initial blog post release.}
\label{fig:external-validation}
\end{figure}

Our optimized nanopath evaluation suite closely tracked performance on these external benchmarks (Figure~\ref{fig:external-validation}; Table~\ref{tab:external-fidelity}). \texttt{nanopath-evals} final score reached Pearson correlations of $r=.927$, $.929$, and $.814$ on combined THUNDER, HEST, and CPTAC, respectively. Using Kendall $\tau$, which assesses how well measurements agree on the rank ordering of models, we observed $.800$, $.789$, and $.684$ on combined THUNDER, HEST, and CPTAC. 

We also compared absolute performance. The strongest nanopath results were $.710$ on combined THUNDER, $.362$ on HEST, and $.657$ on CPTAC, compared with $.761$, $.420$, and $.685$ for the best reference encoders. Despite the large differences in model size and pretraining resources, nanopath models surpassed some large-scale pathology foundation models on a few measures. The \texttt{robust-norm} nanopath model exceeded OpenMidnight 1.0 on THUNDER classification ($.754$ versus $.692$) and H-optimus-0 on segmentation ($.667$ versus $.652$).

For a more detailed comparison, we examined six nanopath checkpoints alongside UNI-2-h, H-optimus-0, Virchow, and DINOv2-G/14. Surprisingly, the \texttt{robust-norm} nanopath model also had the highest BreaKHis macro-F1 under all three THUNDER probes: linear probing ($.841$), $k$-NN ($.821$), and 16-shot SimpleShot ($.772$). Nanopath checkpoints also performed best on six of 11 CPTAC cohort aggregates (see Appendix~\ref{app:detail}). 

\subsection{Limitations}
\label{sec:limitations}
Our external comparison supports nanopath as a low-cost means to experiment with pathology foundation-model training and assess frozen representations. Whether the best small-scale recipes remain best with larger models, more tiles, and longer training schedules remains untested. The optimized \texttt{nanopath-evals} suite was developed after inspecting these same external comparisons, so its correlations measure retrospective agreement and require validation on new encoders and cohorts. Repeated optimization against a public score can induce adaptive overfitting~\citep{cawley2010overfitting}; withholding official test samples from the fast suite does not remove selection effects from using external results to redesign it. The panel includes related checkpoints and different feature readouts. Some evaluation datasets can overlap TCGA pretraining, including PanNuke's mixed-source folds, and UCLA's slide-level folds do not establish patient independence (Appendices~\ref{app:protocol} and~\ref{app:audits}). Selected task wins are point estimates from a restricted comparator set, without pairwise significance tests (Appendix~\ref{app:detail}).

\section{Discussion}
\label{sec:discussion}
Nanopath is built around a simple bet: making pathology foundation-model experiments faster, fairer, and easier to share will help the field discover better training ideas. Much of the structure that makes pathology interesting is difficult to investigate when each hypothesis requires a large pretraining run and time-consuming evaluation. Nanopath gives students, researchers, and AI agents a small, hackable setting in which to try those ideas and get broad feedback quickly. Promising ideas validated with nanopath can then be scaled up to large-scale training runs without constraints.

\texttt{nanopath-evals}, our fast evaluation suite, is also useful on its own, enabling anyone to quickly evaluate pathology foundation models on limited compute. Its strong agreement with standardized benchmarks like THUNDER, HEST, and PathoBench (CPTAC) across 20 encoders, including nanopath models and state-of-the-art pathology foundation models, supports its use as a practical proxy for slower, more comprehensive benchmark evaluations.

Our integration with Labless encourages a more collaborative, crowd-sourced research process. By linking source code, metrics, and hardware for nanopath runs, Labless gives researchers the tools to easily share results and insights with the broader community. Sharing negative results also matters: unsuccessful runs can still narrow a hypothesis, reveal a tradeoff between tasks, or suggest that apparent improvements may have been due to chance. Nanopath's interface was designed to support autoresearch-style experimentation~\citep{karpathy2026autoresearch,medarc2026nanopath}, and many Labless submissions have already been produced this way. Automatic preparation, bounded training, a fixed scalar objective, and our public API-compatible ledger make the loop straightforward to automate. 

The next step is to take promising hypotheses beyond this small-scale setting: matched retraining with larger models and more data, evaluation on stronger patient-separated cohorts, and testing under acquisition shifts and intended clinical use. Community meetings, open code, and donated compute access provide routes for more researchers to participate in that process. We encourage interested readers to join our MedARC Discord server (\url{https://www.medarc.ai/}) to directly collaborate with us on nanopath and its research extensions. 

\section{Author Contributions}
Detailed author contributions are provided in Appendix~\ref{app:author-contributions}.

\section*{Acknowledgments}
We thank Varun Belagali for help setting up efficient HEST and CPTAC standardized benchmarks on our cluster. We thank the TCGA Research Network, specimen donors, and research groups whose data support this work (\url{https://www.cancer.gov/tcga}).

\clearpage
\bibliographystyle{unsrtnat}
\bibliography{references}
\clearpage
\appendix
\input{appendix.tex}
\clearpage
\end{document}

%% file: appendix.tex
\section{Author Contributions}
\label{app:author-contributions}
\textbf{PSS}: conceived the project, led the team, developed nanopath and Labless, designed and optimized the evaluation suite, validated community submissions, ran external benchmark comparisons, analyzed results, prepared figures, and drafted the manuscript.
\textbf{HM}: implemented FINO-style metadata conditioning and shared TCGA metadata, ran training experiments, helped design the evaluation tasks and external comparisons, and contributed to manuscript framing and revisions.
\textbf{RK}: ran extensive autoresearch and ablation experiments across SSL objective composition (DINO/iBOT self-distillation weighting, JEPA latent prediction, KDE and SIGReg~\citep{balestriero2025lejepa} anti-collapse regularizers, FINO metadata-contrastive auxiliaries), stain and local-crop augmentation, masking dose schedules, gradient-conflict projection and weight averaging, and backbone swaps, developed the multi-layer feature readouts and JBU fusion~\citep{kopf2007jbu} used in the \texttt{block-strided-cls} recipe, investigated data curation, patient-level sampling, and reference-model performance, and analyzed seed variability and acquisition robustness.
\textbf{NA}: developed the contiguous-patch I-JEPA recipe, explored masking and augmentation choices, tested nanopath across compute clusters, and shared the implementation and rationale behind the validated recipe.
\textbf{RSG}: contributed feedback on pathology model training, augmentation, and evaluation, informed by related OpenMidnight experiments, and participated in project discussions.
\textbf{SR}: ran and shared exploratory JEPA experiments with molecular metadata conditioning, scanner-adversarial objectives, and changes to hyperparameters and feature readouts.
\textbf{RB}: ran experiments with CLS-token specialization, alternative JEPA predictors, and centering in place of Sinkhorn--Knopp, and shared results and methodological feedback.
\textbf{AD}: developed \texttt{robust-norm} and subsequent feature-readout and test-time augmentation variants, shared the leading recipe and its technical explanation, identified a segmentation image--label alignment issue, and contributed feedback on evaluation and score weighting.
\textbf{CF}: explored alternative tile preparation and sampling, shared a corresponding nanopath run, and contributed feedback on robustness, scale transfer, and the manuscript.
\textbf{CL}: contributed to experimental design, reference-model comparisons, data-loading discussions, and the design of incentives for open collaboration.
\textbf{BW}: increased training throughput by ~150\%, advised on data loading and fair compute constraints, and contributed feedback on evaluation precision, model comparisons, and interpretation.
\textbf{TMA}: oversaw the project, explored alternative self-supervised approaches including DINOv3, CAPI, and SimCLR, and contributed to experimental design, manuscript framing, and revisions.
All authors contributed to research discussions through the Path-FM meetings and community channels.

\section{Reproducibility and evaluation details}
\label{app:protocol}
The paper describes the nanopath release at commit \texttt{92f8d411489a}. The public repository contains the training code, evaluation code, manifests, and benchmark documentation. The evaluation data are packaged under \href{https://huggingface.co/datasets/medarc/nanopath-evals}{\texttt{medarc/nanopath-evals}}. The training tiles are distributed as \href{https://huggingface.co/datasets/medarc/nanopath}{\texttt{medarc/nanopath}}.  

\paragraph{Default nanopath training.}
The default nanopath recipe included in the main GitHub branch initializes DINOv2-S/14 with four registers, uses batch size 128, two global $224\!\times\!224$ views, eight local $98\!\times\!98$ views, and bfloat16 mixed precision training. AdamW uses base learning rate $10^{-4}$, $\beta_2=.99$, layer-wise decay .7, and a patch-embedding learning-rate multiplier of .2. Warmup ends after 30,000 tile presentations. The KDE coefficient is .005 and concentration 10. Schedules other than warmup are keyed to consumed FLOPs; the main recipe reaches the presentation cap before using the full FLOP allowance. Configuration and source are part of each run record. Parameter counts refer to the inference backbone, not training-only teachers and projection heads. Pretraining initialization and one-time data preparation are outside the per-run budget.

\paragraph{Development splits and heads for nanopath-evals.}
Our THUNDER manifest, shared in the nanopath GitHub repo, selects official training and validation data. The 12 classification tasks contain 218--4,096 training and 50--2,048 validation images per task. Every selected class has at least 16 training examples. Capped selections preserve class and relevant source/group structure; SPIDER source slides and WILDS patient/node groups are disjoint. ESCA validation is UKK-only, though the probe-training pool includes TCGA. PanNuke uses all 2,656 Fold1 images for training and all 2,523 Fold2 images for validation, with Fold3 excluded. PanNuke's mixed-source folds include TCGA and local-hospital images without reliable per-image provenance, so Fold2 cannot be certified free of TCGA pretraining overlap. SegPath uses 32,768/4,518 epithelial crops and 20,906/2,164 lymphocyte crops for training/validation, with disjoint source-image sets. Primary classification sources are BACH~\citep{aresta2019bach}, BRACS~\citep{brancati2022bracs}, BreaKHis~\citep{spanhol2016breakhis}, CRC~\citep{kather2018crc}, ESCA~\citep{tolkach2023esca}, MHIST~\citep{wei2021mhist}, PCam~\citep{veeling2018pcam}, SPIDER~\citep{nechaev2025spider}, and WILDS~\citep{koh2021wilds}. PanNuke~\citep{gamper2019pannuke} and SegPath~\citep{komura2023segpath} provide the segmentation data. Source terms are linked in \texttt{benchmarking/classification.md} and \texttt{benchmarking/segmentation.md}.

Linear classification probes average nine Adam heads at learning rates $\{10^{-3},10^{-4},10^{-5}\}$ and weight decays $\{0,10^{-3},10^{-4}\}$, trained for 200 epochs with batch size 64 and seed 0. The $k$-NN score averages $k\in\{1,3,5,10,20,30,40,50\}$. SimpleShot uses centered, normalized features and majority voting over 1,000 support draws. The segmentation head has two layers, eight attention heads, width 192, and feed-forward width 768. Fixed schedules use 30, 9, and 21 epochs for PanNuke, SegPath epithelial, and SegPath lymphocytes; final-epoch weighted per-image macro-F1 is reported. No validation checkpoint or hyperparameter is selected.

For UCLA progression and SurGen mutation~\citep{myles2025surgen}, the prepared \texttt{train}/\texttt{val} storage partitions are pooled into their original PathoBench fold-0 development cohort. Three stratified folds use balanced logistic regression with inverse regularization strength $C=.5$ (distinct from the classification score in Equation~\ref{eq:score}), no feature standardization, and at most 5,000 iterations. UCLA's 90 slide IDs do not establish patient separation: the source study contains repeated samples from patients~\citep{pennycuick2020lung}. Survival probes instead group slides by case, fit the scaler on training cases only, and average CoxNet results at $.1$, $.2$, and $.7$ times the training-fold $\alpha_{\max}$, the smallest penalty yielding all-zero coefficients, with elastic-net mixing $.5$. LEOPARD selects all 87 event cases and the 87 censored cases with longest follow-up from its public training cohort; CPTAC-PDA uses 77 cases with 56 events. These curated development distributions are not population-representative clinical cohorts.

CRoMa measures whether biological similarity dominates acquisition-center similarity~\citep{grisi2026croma}. For each tile, we exclude same-slide neighbors and compare the mean cosine distance to the five nearest same-class/different-center neighbors ($d_{SO}$) with that to five different-class/same-center neighbors ($d_{OS}$). The signed margin is $(d_{OS}-d_{SO})/(d_{OS}+d_{SO})$; the cohort score is its median. Unlike the original fixed-$k$ robustness index, this evaluates every tile with a common support requirement and retains relative distance information. We map each cohort median from $[-1,1]$ to $[0,1]$ and average the two cohorts to obtain $R$.

CRoMa uses PathoROB's prepared cohorts~\citep{komen2026pathorob}: 22,402 Camelyon tiles from 97 slides~\citep{litjens2018camelyon} and 13,800 Tolkach ESCA tiles from 62 slides~\citep{tolkach2023esca}. Tolkach's 2,500 TCGA-center tiles are excluded. The fixed representation concatenates final normalized CLS and mean patch tokens, followed by L2 normalization. Cosine-distance arithmetic is float64. The implementation matches upstream \texttt{clemsgrs/croma} commit \texttt{3f58d5e4bd9e}; sample margins agreed within $1.06\!\times\!10^{-11}$ on 58 deterministic cohort/model checks. Besides the median, the output retains lower-tail margins and the fraction of non-positive margins, because a favorable median can hide vulnerable subgroups.

\section{Original suite and recipe records}
\label{app:original}
The original suite used four classification datasets and three nucleus-segmentation datasets (PanNuke, MoNuSAC~\citep{verma2021monusac}, and CoNSeP~\citep{graham2019hovernet}), plus the slide and robustness cohorts in Table~\ref{tab:nanopath_eval_v1}. Its score was
\begin{equation}
S_{\mathrm{original}}=\frac{C_{\mathrm{linear}}+C_{k\mathrm{NN}}+C_{\mathrm{shot}}+\mathrm{Seg}+P+M+\mathrm{Sur}+R_{\mathrm{RI}}}{8}.
\end{equation}
Here $S_{\mathrm{original}}$ is the original final score; $C_{\mathrm{linear}}$, $C_{k\mathrm{NN}}$, and $C_{\mathrm{shot}}$ are dataset-mean classification macro-F1 for linear, $k$-NN, and 16-shot probes. $\mathrm{Seg}$ is mean segmentation macro-Jaccard; $P$, $M$, and $\mathrm{Sur}$ denote the progression, mutation, and survival family means; and $R_{\mathrm{RI}}$ is mean PathoROB robustness index. Unlike the optimized suite, some probe hyperparameters were selected on the validation observations being reported. Original and optimized scores therefore cannot be compared as if they used the same scale or protocol.

\begin{table}[t]
\centering
\caption{Summary of the original nanopath evaluation suite.}
\label{tab:nanopath_eval_v1}
\small
\setlength{\tabcolsep}{5pt}
\renewcommand{\arraystretch}{1.05}
\begin{tabularx}{\textwidth}{@{}lXll@{}}
\toprule
Family & Datasets & Probe & Metric \\
\midrule
Classification
& BreaKHis, BRACS, MHIST, PCam
& Linear, $k$-NN, 16-shot SimpleShot
& Macro-F1 \\
Segmentation
& PanNuke, MoNuSAC, CoNSeP
& Frozen MaskTransformer
& Macro-Jaccard \\
Progression
& UCLA Lung
& Mean pooling + logistic regression
& Macro-OVR AUROC \\
Mutation
& SurGen RAS
& Mean pooling + logistic regression
& Macro-OVR AUROC \\
Survival
& LEOPARD BCR, CPTAC-PDA OS
& Mean pooling + ridge CoxPH
& Harrell's c-index \\
Robustness
& PathoROB Camelyon, Tolkach ESCA
& Different-slide $k$-NN
& Robustness index \\
\bottomrule
\end{tabularx}
\end{table}

\subsection{Exact recipe records}
\label{app:lineage}
Table~\ref{tab:runids} identifies the six checkpoint records and the training durations used in Table~\ref{tab:validated-lineage}. The archived ledger distinguishes reviewed or validated reruns from promoted leaders. In particular, the \texttt{jepa-fino} record explicitly states that its seed-2026 rerun did not clear the 0.006 margin. Later rows should not be read as single-factor ablations: the I-JEPA recipe does not retain all curation/optimizer changes from the preceding row, and several recipes change both learning and feature readout.

\begin{table}[ht]
\centering
\caption{Public Labless records used for the historical recipe table. Identifiers link directly to the corresponding source-linked record. Times are reported by the recorded run, not standardized timing repetitions.}
\label{tab:runids}
\small
\begin{tabular}{@{}llr@{}}
\toprule
Recipe & Labless run identifier & Training seconds \\
\midrule
dinov2-s-kde & \href{https://labless.dev/runs/run_sub_0d8aeb2511}{\texttt{run\_sub\_0d8aeb2511}} & 4381.5 \\
lr-and-curation & \href{https://labless.dev/runs/run_sub_6c6c051f71}{\texttt{run\_sub\_6c6c051f71}} & 3550.8 \\
i-jepa-contig-patch & \href{https://labless.dev/runs/run_sub_816fc5d0ca}{\texttt{run\_sub\_816fc5d0ca}} & 3472.3 \\
jepa-fino-s2026 & \href{https://labless.dev/runs/run_sub_1879d32919}{\texttt{run\_sub\_1879d32919}} & 3901.6 \\
block-strided-cls & \href{https://labless.dev/runs/run_sub_59d24d6b7b}{\texttt{run\_sub\_59d24d6b7b}} & 3265.4 \\
robust-norm-s9876 & \href{https://labless.dev/runs/run_sub_16b156161d}{\texttt{run\_sub\_16b156161d}} & 6041.2 \\
\bottomrule
\end{tabular}
\end{table}

\section{External benchmark protocol and full model panel}
\label{app:external}
Our internal benchmark relies on code from the official THUNDER, HEST, and PathoBench codebases. Optimizations include fp16 encoder inference, cached features, vectorized frozen probes, int8 dense-token storage with fp16 scales, and TF32 segmentation-decoder matrix multiplication. They retain benchmark tasks and splits but are not bitwise executions of upstream code. The public protocol is available at \url{https://github.com/MedARC-AI/pathfm-full-evals}.

THUNDER classification averages all 16 task datasets for each of linear, $k$-NN, and 16-shot probing; segmentation averages four tasks, including TCGA-derived OCELOT. The combined target is their equal-weight mean, not THUNDER's six-column rank-sum leaderboard. HEST retains the nine official tasks, patient-level splits, PCA-256, ridge regression, and 50-gene scoring. CPTAC retains 38 classification tasks across 11 cohorts and official splits, excluding the separate combined-organ task. Internal point estimates do not include the full upstream THUNDER bootstrap analysis. Published and internally reproduced scores are combined at the exact-checkpoint level; H0-mini's THUNDER segmentation value, for example, is the published score.

The reference panel includes natural-image DINOv2-S/B/L/G~\citep{oquab2023dinov2}; UNI-2-h~\citep{chen2024uni}, Virchow~\citep{vorontsov2024virchow}, GigaPath~\citep{xu2024gigapath}, Midnight-12K~\citep{karasikov2025midnight}, and OpenMidnight~\citep{openmidnight2025}; and H-optimus-0, H0-mini, GenBio-PathFM, Kaiko-S/16, and GigaPath-Flash~\citep{saillard2024hoptimus,filiot2025h0mini,kapse2026genbiopathfm,kaiko2024largescale,usuyama2026gigapathflash}. These frozen reference encoders are comparison points, not training-budget-matched submissions. In particular, small distilled references inherit information from larger pathology teachers; parameter count alone does not measure pretraining exposure.

For the optimized fast suite, clean-process H100 evaluations ranged from 1,018 to 1,366 seconds for the documented reference/readout checks. After CRoMa was introduced, the evaluation phases of complete retraining checks took 1,146.8 seconds for main and 1,135.4 seconds for \texttt{robust-norm}. Full external evaluation schedules at most two concurrent one-H100 jobs per checkpoint and separate CPU fits. Representative THUNDER stages take 20--30 minutes for classification feature extraction, 1.75--2.25 hours for cached heads, and 4--4.5 hours for segmentation plus attacks; CPTAC extraction takes 50--60 minutes. 

\begin{table}[ht]
\centering
\caption{Frozen 20-model comparison underlying the correlation analysis. $S_1/S_2$ are original/optimized nanopath scores. External columns are THUNDER classification (Cls), segmentation (Seg), HEST mean expression correlation, and CPTAC mean AUROC. All are point estimates; linked model names identify their public sources.}
\label{tab:allmodels}
\small
\setlength{\tabcolsep}{4pt}
\begin{tabularx}{\textwidth}{@{}Xrrrrrr@{}}
\toprule
Encoder & $S_1$ & $S_2$ & Cls & Seg & HEST & CPTAC \\
\midrule
dinov2-s-kde & 0.628 & 0.627 & 0.717 & 0.656 & 0.339 & 0.646 \\
lr-and-curation & 0.636 & 0.633 & 0.719 & 0.655 & 0.353 & 0.643 \\
i-jepa-contig-patch & 0.644 & 0.641 & 0.734 & 0.651 & 0.356 & 0.656 \\
block-strided-cls & 0.659 & 0.639 & 0.748 & 0.663 & 0.362 & 0.650 \\
robust-norm & 0.668 & 0.648 & 0.754 & 0.667 & 0.356 & 0.650 \\
jepa-fino & 0.648 & 0.646 & 0.745 & 0.657 & 0.354 & 0.657 \\
\midrule
\href{https://huggingface.co/genbio-ai/genbio-pathfm}{GenBio-PathFM} & 0.692 & 0.700 & 0.827 & 0.682 & 0.420 & 0.685 \\
\href{https://huggingface.co/MahmoodLab/UNI2-h}{UNI-2-h} & 0.678 & 0.694 & 0.829 & 0.693 & 0.414 & 0.675 \\
\href{https://huggingface.co/bioptimus/H-optimus-0}{H-optimus-0} & 0.676 & 0.683 & 0.805 & 0.652 & 0.415 & 0.673 \\
\href{https://huggingface.co/paige-ai/Virchow}{Virchow} & 0.659 & 0.678 & 0.773 & 0.691 & 0.406 & 0.661 \\
\href{https://huggingface.co/prov-gigapath/prov-gigapath}{GigaPath} & 0.646 & 0.668 & 0.793 & 0.671 & 0.388 & 0.672 \\
\href{https://huggingface.co/kaiko-ai/midnight}{Midnight-12K} & 0.620 & 0.694 & 0.787 & 0.648 & 0.395 & 0.648 \\
\href{https://sophont.med/blog/openmidnight/}{OpenMidnight} & 0.611 & 0.638 & 0.692 & 0.693 & 0.391 & 0.656 \\
\href{https://huggingface.co/bioptimus/H0-mini}{H0-mini} & 0.664 & 0.681 & 0.795 & 0.691 & 0.403 & 0.678 \\
DINOv2-S/14 & 0.584 & 0.594 & 0.687 & 0.637 & 0.324 & 0.635 \\
DINOv2-B/14 & 0.595 & 0.599 & 0.689 & 0.639 & 0.334 & 0.643 \\
DINOv2-L/14 & 0.612 & 0.612 & 0.698 & 0.642 & 0.335 & 0.638 \\
DINOv2-G/14 & 0.620 & 0.615 & 0.711 & 0.642 & 0.343 & 0.643 \\
\href{https://github.com/kaiko-ai/towards_large_pathology_fms}{Kaiko-S/16} & 0.629 & 0.641 & 0.784 & 0.668 & 0.349 & 0.637 \\
\href{https://huggingface.co/prov-gigapath/prov-gigapath-flash}{GigaPath-Flash} & 0.642 & 0.650 & 0.771 & 0.620 & 0.362 & 0.659 \\
\bottomrule
\end{tabularx}
\end{table}

\begin{table}[ht]
\centering
\caption{Original versus optimized proxy fidelity on the same 20 encoders. Entries are Pearson $r$/Kendall $\tau$. The first two rows compare matching family scores; the final three use the overall nanopath score. Figure~\ref{fig:external-validation} shows both versions for the four individual targets; the combined THUNDER target is the equal-weight mean of classification and segmentation.}
\label{tab:external-fidelity}
\small
\begin{tabularx}{\textwidth}{@{}Xcc@{}}
\toprule
External target & Original & Optimized \\
\midrule
THUNDER classification & $.812/.663$ & $.988/.958$ \\
THUNDER segmentation & $.626/.508$ & $.870/.741$ \\
THUNDER classification + segmentation & $.829/.695$ & $.927/.800$ \\
HEST gene expression & $.718/.642$ & $.929/.789$ \\
CPTAC classification & $.792/.600$ & $.814/.684$ \\
\bottomrule
\end{tabularx}
\end{table}

\section{Seed, null, and score-policy audits}
\label{app:audits}
We repeatedly trained the main (lr-and-curation) and robust-norm nanopath models using three random seeds with data splits and probe randomness fixed. Under the optimized score, their final-score sample standard deviations were .002337 and .000438, respectively. The pooled within-recipe standard deviation was .001681. These six runs inform our use of a fixed .004 threshold for reliable run improvements. 

\begin{table}[ht]
\centering
\caption{Training-seed sensitivity under the current score. Each cell is a separately trained checkpoint with the same fixed downstream protocol.}
\small
\begin{tabular}{@{}lrrrrr@{}}
\toprule
Recipe & Seed 17 & Seed 29 & Seed 43 & Mean & Sample SD \\
\midrule
Main & .619722 & .620240 & .615958 & .618640 & .002337 \\
Robust-norm & .645294 & .645239 & .646023 & .645518 & .000438 \\
\bottomrule
\end{tabular}
\end{table}

A separate ten-seed audit randomized DINOv2-S backbones while retaining the production probes. The optimized final score was $.4731\pm.0028$ (mean $\pm$ sample standard deviation), range $.4682$--$.4764$. Every trained/pretrained encoder in Table~\ref{tab:allmodels} exceeds that maximum by at least .118. This aggregate separation coexists with weak individual probes (Table~\ref{tab:null}). Above-chance random features may reflect useful low-level image statistics, cohort shortcuts, or weaknesses in the evaluation unit; these explanations have not been separated. In particular, UCLA's slide-level folds cannot certify patient independence.

\begin{table}[ht]
\centering
\caption{Random-feature audit of ten independently initialized encoders. The final row rescales signed CRoMa to $[0,1]$ as in Equation~\ref{eq:score}.}
\label{tab:null}
\small
\begin{tabular}{@{}lrrr@{}}
\toprule
Component & Mean & Sample SD & Range \\
\midrule
Classification & .3706 & .0026 & .3661--.3739 \\
Segmentation & .5128 & .0047 & .5067--.5202 \\
Progression & .6684 & .0085 & .6576--.6841 \\
Mutation & .5502 & .0038 & .5437--.5558 \\
Survival & .5985 & .0076 & .5842--.6100 \\
Robustness & .2706 & .0071 & .2569--.2795 \\
\bottomrule
\end{tabular}
\end{table}

\section{Scope of the task-level comparisons}
\label{app:detail}
Table~\ref{tab:taskwins} supplies the individual-task evidence highlighted in the main text. The comparator set is restricted to the four displayed references; the six nanopath checkpoints were searched for the largest point estimate in each row. Selecting the best nanopath checkpoint per task introduces selection effects, and no pairwise significance claim is made.

\begin{table}[ht]
\centering
\caption{Selected task-level point estimates in the ten-model comparison. These rows substantiate the specific wins discussed in the text; they are not an unbiased summary across tasks.}
\label{tab:taskwins}
\small
\setlength{\tabcolsep}{3pt}
\begin{tabularx}{\textwidth}{@{}lXrrrrr@{}}
\toprule
Task & Best nanopath & Score & UNI-2-h & H-opt.-0 & Virchow & DINOv2-G \\
\midrule
BreaKHis linear & robust-norm & 0.841 & 0.784 & 0.687 & 0.627 & 0.809 \\
BreaKHis k-NN & robust-norm & 0.821 & 0.794 & 0.819 & 0.560 & 0.762 \\
BreaKHis 16-shot & robust-norm & 0.772 & 0.699 & 0.706 & 0.505 & 0.614 \\
CPTAC BRCA & i-jepa-contig-patch & 0.692 & 0.684 & 0.688 & 0.690 & 0.669 \\
CPTAC CCRCC & i-jepa-contig-patch & 0.579 & 0.570 & 0.569 & 0.551 & 0.557 \\
CPTAC GBM & i-jepa-contig-patch & 0.715 & 0.709 & 0.689 & 0.677 & 0.708 \\
CPTAC HNSC & lr-and-curation & 0.633 & 0.630 & 0.617 & 0.612 & 0.569 \\
CPTAC OV & robust-norm & 0.602 & 0.535 & 0.540 & 0.512 & 0.533 \\
CPTAC PDA & i-jepa-contig-patch & 0.592 & 0.469 & 0.516 & 0.564 & 0.565 \\
\bottomrule
\end{tabularx}
\end{table}

\section{Open research and responsible use}
\label{app:impact}
Nanopath is released under Apache-2.0. The code, manifests, and benchmark documentation credit the upstream projects and link their original data terms; the evaluation mirror does not relicense the source datasets. DINOv2 initialization weights and reference checkpoints retain their respective licenses. Researchers must meet those access and use conditions. The study reuses existing pathology datasets and public experiment records; it does not recruit patients or conduct a clinical intervention.

MedARC's public research meetings and community provide a place to discuss experiments, and our launch announcement describes how we donate compute access for volunteers~\citep{nanopath2026blog}. These mechanisms lower practical barriers to participation. 

Potential harms include overinterpreting benchmark gains as clinical readiness, perpetuating acquisition or cohort biases, and rewarding repeated optimization of a weak proxy. Versioned protocols, source inspection, independent reruns, null audits, and explicit data-overlap disclosures help expose these risks. The reported models have not been validated for clinical decision-making. 